# Analysis of different temporal graph neural network configurations on dynamic graphs

Project Report - IST597 Foundations of Deep Learning

**Ashmita Bhattacharya**
Department - CEE
Pennsylvania State University
State College, PA, USA
azb6481@psu.edu

**Rishu Verma**
Department - EECS
Pennsylvania State University
State College, PA, USA
rfv5129@psu.edu

**Sai Naveen Katla**
Department - EECS
Pennsylvania State University
State College, PA, USA
svk6448@psu.edu

**PROBLEM STATEMENT**

In recent years, there has been an increasing interest in the use of graph neural networks (GNNs) for analyzing dynamic graphs, which are graphs that evolve over time. However, there is still a lack of understanding of how different temporal graph neural network (TGNs) configurations can impact the accuracy of predictions on dynamic graphs. Moreover, the hunt for benchmark datasets for these TGNs models is still ongoing. Up until recently, Pytorch Geometric Temporal came up with a few benchmark datasets but most of these datasets have not been analyzed with different TGN models to establish the state-of-the-art. Therefore, this project aims to address this gap in the literature by performing a qualitative analysis of spatial-temporal dependence structure learning on dynamic graphs, as well as a comparative study of the effectiveness of selected TGNs on node and edge prediction tasks. Additionally, an extensive ablation study will be conducted on different variants of the best-performing TGN to identify the key factors contributing to its performance. By achieving these objectives, this project will provide valuable insights into the design and optimization of TGNs for dynamic graph analysis, with potential applications in areas such as disease spread prediction, social network analysis, traffic prediction, and more.

## 1 INTRODUCTION

**Graphs** are used to represent complex relationships between entities, thereby helping provide a flexible and intuitive way to model a complex problem in a neat visual representation. The idea of representing data as a graph has been around for a long time and has been used in various fields such as social network analysis, traffic flow, network modeling, natural language processing, etc. With the introduction of the Graph Convolutional Network (GCN) architecture in around 2016 [7]. the attention diverted to using graph-structured data and since then, the research on GNNs has exploded, and various **GNN** architectures have been proposed, including Graph Attention Networks (GAT), GraphSAGE, and Message Passing Neural Networks (MPNNs), etc.

These traditional GNNs operate on **static graphs**, where the graph structure and node features are assumed to be fixed in time. However, in many real-world applications, the graph structure and node features can change over time - **dynamic graphs**. For example, in social media networks, the connections between users can change over time and the same goes for problems like traffic flow prediction, disease spread prediction, etc. In order to analyze these dynamic graphs, **Temporal Graph Neural Network (TGNs)** architecture was introduced as an extension of the Graph Neural Network (GNN) model. In TGNs[4] the graph structure includes an additional dimension of time series (aka temporal signals) along with the spatial dimension (which is already considered in traditional GNN) and the model is trained to predict the future states of the graph.

With this excitement and motivation around TGNs, we performed a qualitative analysis on a few of the significant TGN models namely, **EvolveGCN** (Evolve Graph Convolutional Networks), **GConv-LSTM**, and **GConv-GRU** on a node-level regression problem of disease spread prediction where graphs are represented as a sequence of snapshots. On the other hand, other significant event-based models like **TGAT** (Temporal Graph Attention Networks), and **TGN** (Temporal Graph Networks), which process graphical data as a sequence of events are also studied in an edge-level prediction task. Further details regarding the project are organized in the upcoming sections of the report: Section 2 provides a review of relevant literature, Section 3 describes the methodology and datasets used in the project, Section 4 presents the results of the experiments, Section 5 & 6 discusses the ablation study and challenges respectively, Section 7 focus on the future work, Section 8 concludes the report with a brief summary.

## 2 RELEVANT LITERATURE

This section provides a comprehensive overview of the prior work in the field and a brief introduction to the TGN models we intend to implement as part of our project. A brief discussion will also be provided about the motivation behind choosing the dataset.

Following the taxonomy introduced in [6], the different existing TGNN models can be broadly classified into two sections- Snapshot-based or Discrete Time Dynamic

Graphs (DTDG) and Event-based or Continuous Time Dynamic Graphs (CTDG). Snapshot-based models are used to learn over sequences of time-stamped static graphs. These methods are suitable to process the entire graph at each point in time with a suitable mechanism to aggregate temporal dependencies across different time-steps. On the other hand, event-based models process the time-evolving graphs as sequences of events and update a node or/and an edge if an event occurs involving that node or edge.

## 2.1 Snapshot-based Models

Snapshot-based models or DTDGs can be further distinguished as model-evolution and embedding-evolution based models. In evolution-based models, the parameters of static GNN evolve over time. EvolveGCN [1] is an example of this category of snapshot-based models, which is described in detail in 2.1.1. Embedding-evolution based models learn on embedding space (instead of parameter space) learned by static models. There are several models which fall under this category and we have selected DySAT [2] detailed in 2.1.2.

### 2.1.1 EvolveGCN[1]

EvolveGCN extends the Graph Convolutional Network (GCN) architecture to dynamic graphs by introducing a temporal attention mechanism that adaptively combines information from multiple time steps. The algorithm also uses a gating mechanism to control the flow of information between the current and past states of the graph. In [1] the author has evaluated EvolveGCN mostly on the undirected graph datasets like Reddit, SBM, Bitcoin-OTC/Alpha, UCI, AS, and Elliptic and shows that it outperforms existing approaches to modeling dynamic graphs for tasks like link prediction, edge classification, and node classification.

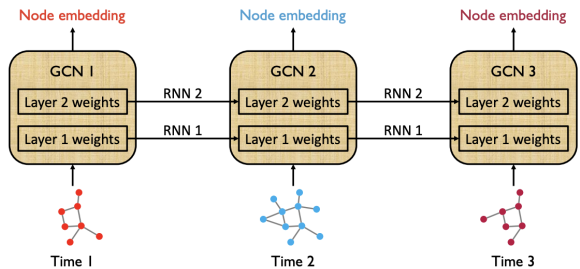

[1] Figure 1: Schematic illustration of EvolveGCN. The RNN means a recurrent architecture in general (e.g., GRU, LSTM).

In the paper, the author has introduced two versions of EvolveGCN namely the -O and -H versions. In the -O version, the weights are treated as input and output from the LSTM (recurrent architecture) but in the case of the -H version (the H is inspired by hidden states), the weights are treated as hidden states to the GRU (recurrent architecture).

### 2.1.2 DySAT[2]

Dynamic Self-Attention Network (DySAT) is a neural network model built on a self-attention mechanism that allows nodes in the graph to aggregate information from other nodes based on their feature similarity. Figure 3 shows the architecture of DySAT showing at first the self-attention mechanism is used to generate static node embeddings at each timestamp. Then self-attention blocks are introduced to process past temporal embeddings for a node to generate the novel embedding (or predicted embeddings). In [2] the author has evaluated the model on communication networks (UCI and Enron) and bipartite rating networks (Yelp and MovieLens).

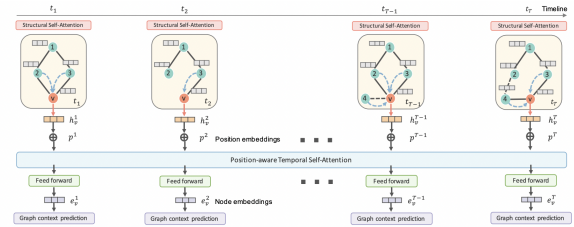

[2] Figure 2: Neural architecture of DySAT showing structural attention layers followed by temporal attention layers

## 2.2 Event-based Models

Event-based models or CTDGs can be further distinguished as temporal-embedding and temporal-neighborhood based models. Temporal-embedding based models learn temporal signals by aggregating information based on time embeddings, and features and connectivity structure of the graph. TGAT[3] is an example of this category, which is described in detail in 2.2.1. Temporal-neighborhood based models use a "mailbox" module to store functions of events to update node/edge embeddings with time based on past events. TGN[4] is chosen here as an example of this type of model, which usually achieves state-of-the-art performance on certain temporal graphs.

### 2.2.1 TGAT[3]

This work involves developing a functional mapping from continuous time to vector time encoding for each node. Using this translation-invariant explicit functional time encoding based on Random Fourier Features (RFF), TGAT introduces graph temporal attention mechanism to aggregate the embeddings of the temporal neighborhoods of a node, where the positional encoding is replaced by the temporal encodings which are learned. The TGAT layer can be thought of as a local aggregation operator that takes the temporal neighborhood with their hidden representations (or features), as well as timestamps as input, and the output, is the time-aware representation for the target node at any time point t. The model is run on three datasets- Wikipedia, Reddit, and industrial on both transductive and inductive tasks.

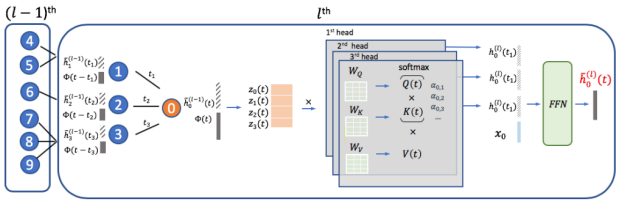

[3] Figure 3: The architecture of l-th TGAT layer at time t for node v0

## 2.2.2 TGN[4]

TGN consists of an encoder that creates compressed representations of nodes based on their interactions and updates them based on each event. Each node seen by the model so far is characterized by a memory vector, which stores a function of its past interactions. Given a new event, a mailbox module computes mail for every node involved. Mails are used to update the memory vector. To overcome the so-called staleness problem, an embedding module also computes, at each time step, the node embeddings using their neighborhoods and memory states.

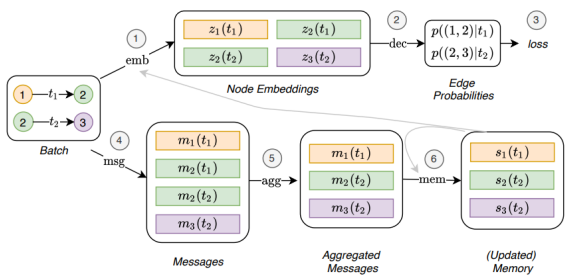

[4] Figure 4: TGN computation on a batch of time-stamped samples

## 3 DATASETS

**EnglandCovid-** We make use of the EnglandCovid Dataset [9] this dataset to understand the continuous dynamics of COVID-19 focusing on dates between 13th March 2020 to 12th May 2020, i.e, 61 days of data. The focus is to forecast the spread of the disease in England NUTS3 regions [8]. The data was collected from smartphones with Facebook installed with location history enabled. The raw data that was collected had data points from three times of the day – morning, afternoon, and evening; to make the dataset efficient for usage and analysis we aggregated the three data points into one single point for that entire day.

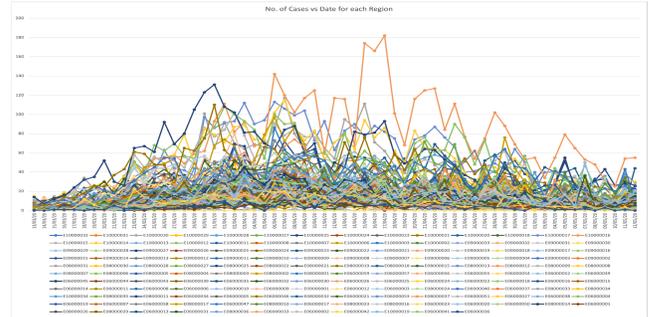

Figure 5: Number of Covid-19 cases with time across different regions under case study.

The data quantified people who travel across the region on that particular day at some given time. The single value in each cell of the dataset is indicating aggregated reading (of all three times of the day) for each pair of regions. In-depth, the name (i.e., E10000031 to E10000032) relates to one pair of regions across which the number of people is moving about, and the number under each date is for the aggregated number of people.

The dataset for England specifically consists of 129 regions for a total of 61 days. Each region is represented as a node in the graph and each edge represents the movement of people from one region to another. Edge attributes (weights) represent the number of movements between the two regions. Figure 6 visualizes the dataset for the first snapshot. Since this dataset intuitively falls under the category of DTDGs where the data over the entire graph is available at regular intervals of time. Therefore, we have used this dataset to study the snapshot-based models.

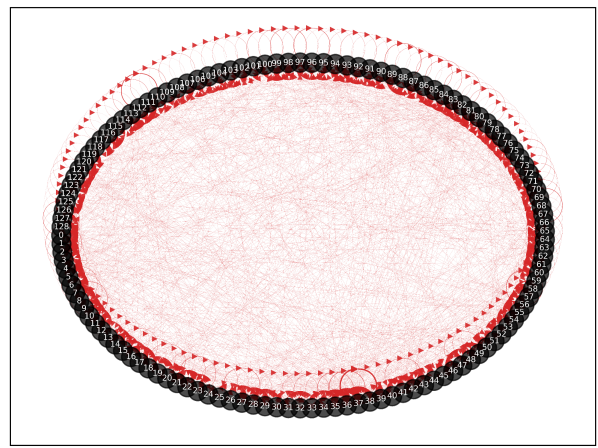

Figure 6: Representation of EnglandCovid Dataset for snapshot $t_0$

# IST-597 Project Report - Team 14

**Wikipedia-** In order to study CTDG models, we select the Wikipedia dataset where an event representing any node or edge-related activity can happen at any continuous time. The dataset consists of 30 days of edits made by users on Wiki pages after selecting the 1000 most edited pages and users who have edited at least 5 pages resulting in a total of 8227 users. Thus, the total number of nodes is 9227 (8227 users and 1000 pages), and the total number of interactions or edges occurring over time is 157,474. The features of each interaction are obtained by converting the text in each edit made into an LIWC-feature vector of 172 dimensions. The dataset is available at [12].

We focus on an edge-level task on this dataset which is to predict a future interaction, that is given all the interactions till time t, the task is to predict the probability of a user *u* to editing a page *i* at time t. This task is self-supervised.

Since this dataset naturally consists of events happening at irregular intervals of time, this is tackled with the CTDG models like TGAT and TGN. The performance is also compared against baselines like Jodie [11] and DyRep [13] which are also CTDG models.

## 4 RESULTS

### 4.1 ANALYSIS OF SNAPSHOT-BASED MODEL

The dataset (EnglandCovid) is split into training and test as 80% and 20% respectively. Node Features is the normalized value of targets for previous "lag or time step" days, where time step/lag is the input from the user (for the baseline model we have fixed lags as 8). The models are trained for 200 epochs. The idea is to use the node features (which is the number of positive cases over **i+lag** number of days, where i = iterator over the snapshot for 129 regions) to predict the number of positive cases in 129 regions for (i+lag)+1$^{st}$ day.

As part of the comparative study, we observed the mean square error (MSE) trend as follows:

| Model Name | MSE | | |
|---|---|---|---|
| | Time_Step/lag = 4 | Time_Step/lag = 6 | Time_Step/lag = 8 |
| Graph Convolutional Long Short Term Memory (gconv_LSTM) | 0.5243 | 0.6746 | 0.7971 |
| Graph Convolutional Gated Recurrent Unit (gconv_gru) | 0.8713 | 0.9092 | 0.9290 |
| Evolving Graph Convolutional without Hidden Layer (EvolveGCNO) | 0.8189 | 0.8699 | 0.9068 |
| Evolving Graph Convolutional with Hidden Layer (EvolveGCNH) | 0.8375 | 0.8787 | 1.0987 |

Table 4.1.1: MSE of snapshot-based models with varying step sizes.

It is observed that the MSE for gconv_LSTM is the lowest and shows the best performance when we decrease the number of lags (past windows) by 50%. Another observation is that the MSE is the highest for EvolveGCNH. Therefore in order to get better performance we changed the hyperparameters of the models (while keeping the lags = 8) and the tables 4.1.1, 4.1.2, 4.1.3, and 4.1.4 show the observations on each model.

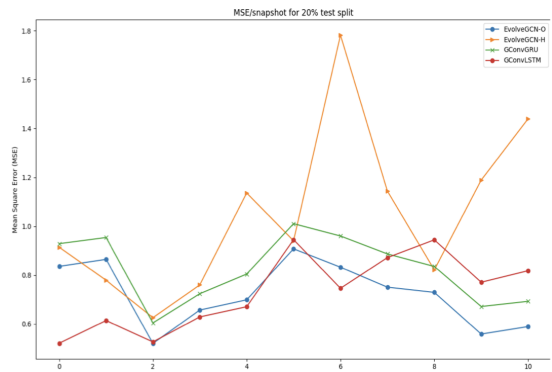

Figure 7: MSE per snapshot on baseline configuration.
**(Baseline configuration: Activation function = ReLU, Learning Rate = 0.01, Optimizer = ADAM, Lags = 8)

| Learning Rate | Activation Function | Optimizer | MSE |
|---|---|---|---|
| 0.1 | ReLU | Adam | 0.883 |
| 0.01 | ReLU | Adam | 0.943 |
| 0.001 | ReLU | Adam | 0.917 |
| 0.0001 | ReLU | Adam | 0.701 |
| 0.1 | tanh | Adam | 0.935 |
| 0.01 | tanh | Adam | 0.914 |
| 0.001 | tanh | Adam | 0.9 |
| 0.0001 | tanh | Adam | 0.616 |
| 0.0001 | tanh | SGD | 0.808 |
| 0.0001 | tanh | SGDM(0.5) | 1.041 |
| 0.0001 | tanh | SGDM(0.9) | 0.838 |
| **0.0001** | **tanh** | **RMSProp** | **0.6109** |

Table 4.1.2: Analysis on GConvGRU

| Learning Rate | Activation Function | Optimizer | MSE |
|---|---|---|---|
| 0.1 | ReLU | Adam | 0.744 |
| 0.01 | ReLU | Adam | 0.734 |
| 0.001 | ReLU | Adam | 0.836 |
| 0.0001 | ReLU | Adam | 0.799 |
| **0.1** | **tanh** | **Adam** | **0.634** |
| 0.01 | tanh | Adam | 0.716 |
| 0.001 | tanh | Adam | 0.849 |
| 0.0001 | tanh | Adam | 0.802 |
| 0.1 | tanh | SGD | 0.967 |
| 0.1 | tanh | SGDM(0.5) | 1.063 |
| 0.1 | tanh | SGDM(0.9) | 0.673 |
| 0.1 | tanh | RMSProp | 0.6407 |

Table 4.1.3: Analysis on GConvLSTM

| Learning Rate | Activation Function | Optimizer | MSE |
|---|---|---|---|
| 0.1 | ReLU | Adam | 0.889 |
| 0.01 | ReLU | Adam | 0.976 |
| 0.001 | ReLU | Adam | 0.865 |
| 0.0001 | ReLU | Adam | 0.735 |
| 0.1 | tanh | Adam | 0.924 |
| 0.01 | tanh | Adam | 0.886 |
| 0.001 | tanh | Adam | 0.9 |
| **0.0001** | **tanh** | **Adam** | **0.538** |
| 0.0001 | tanh | SGD | 0.586 |
| 0.0001 | tanh | SGDM(0.5) | 1.171 |
| 0.0001 | tanh | SGDM(0.9) | 0.84 |
| 0.0001 | tanh | RMSProp | 0.6779 |

Table 4.1.4: Analysis on EvolveGCN-O

| Learning Rate | Activation Function | Optimizer | MSE |
|---|---|---|---|
| 0.1 | ReLU | Adam | 0.959 |
| 0.01 | ReLU | Adam | 0.979 |
| 0.001 | ReLU | Adam | 1.118 |
| 0.0001 | ReLU | Adam | 1.194 |
| 0.1 | tanh | Adam | 1.351 |
| 0.01 | tanh | Adam | 1.275 |
| 0.001 | tanh | Adam | 0.808 |
| 0.0001 | tanh | Adam | 0.841 |
| 0.001 | tanh | SGD | 1.037 |
| **0.001** | **tanh** | **SGDM(0.5)** | **0.775** |
| 0.001 | tanh | SGDM(0.9) | 1.119 |
| 0.001 | tanh | RMSProp | 0.8834 |

Table 4.1.5: Analysis on EvolveGCN-H

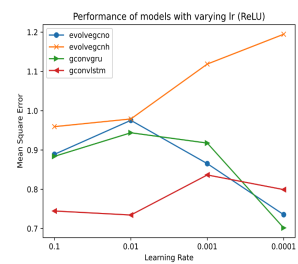
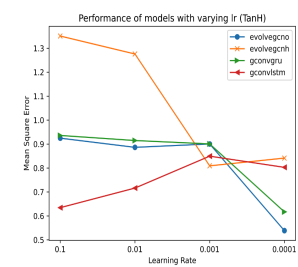

### 4.2 ANALYSIS OF EVENT-BASED MODEL

**Wikipedia-** Two models which are mainly implemented on the edge prediction task on this dataset are TGN and TGAT. TGN is seen to outperform other models on this dataset, and detailed ablation studies shown in section 5 have been performed on TGN to better understand the key components leading to its superior performance.

As shown in Figure 8, the training of TGN is fairly quick with validation precision reaching sufficiently high values during the early stages of training. As seen in Table 2. TGN with an attention mechanism to construct the embedding of each event performs much better on transductive and inductive (evaluation on previously unseen nodes during training) tasks than the other models.

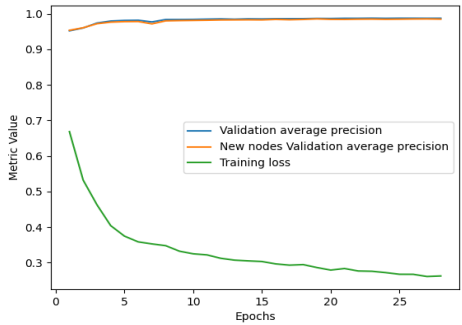

Figure 8: Training progress of TGN.

|  | Test ap | | Test auc | |
| --- | --- | --- | --- | --- |
| Model | Seen | Unseen | Seen | Unseen |
| **TGN-attn** | **0.9858** | **0.9801** | **0.9852** | **0.9753** |
| TGAT | 0.9334 | 0.9021 | 0.9283 | 0.8949 |
| Jodie | 0.9597 | 0.9427 | 0.9566 | 0.9433 |
| DyRep | 0.9437 | 0.9211 | 0.9388 | 0.9118 |

Table 4.2.1: TGN with attention shows the best performance. The other models tested include TGAT which can be shown to be a specific case of TGN, Jodie, and DyRep.

## 4.3 TGN on EnglandCovid Dataset

In order to run the TGN & TGAT on EnglandCovid Dataset the dataset had to be converted from snapshot to event-based data. But at the time of this project, the community is still looking for ways to convert snapshots to the stream of events or Temporal Data. Therefore, we tried to convert the discrete-time dynamic graph (DTDG) to a continuous-time dynamic graph (CTDG) and adjust it to the model's compatibility. As per the structure of the dataset, the first task was to identify the events which could be Node addition and deletion, Edge addition and deletion, and change in edge or node attributes. The second task was to identify the relevant features which could be movements and past cases.

Figure 9 shows the structure of the EnglandCovid Dataset from which (after several trials and errors) we identify events at each timestamp as the number of edges added at that particular timestamp with features as the movements and past cases in the source and destination. More details on the dataset can be found in the GitHub repo [10].

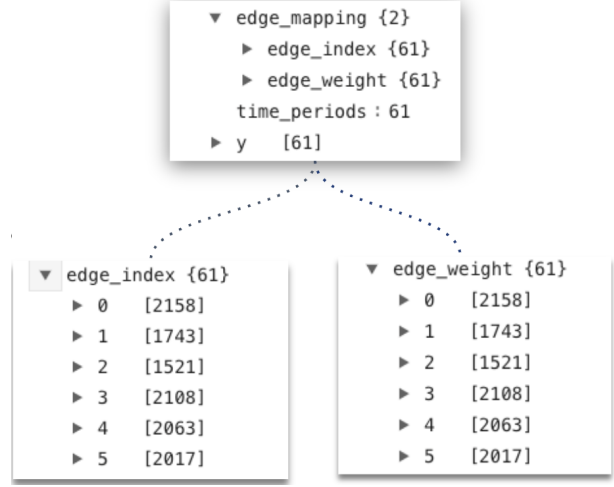

Figure 9: Structure of EnglandCovid Dataset

This converted dataset gives the following results of the edge prediction and node regression task on TGN.

| **Edge Prediction Task** | |
| --- | --- |
| Old nodes test accuracy | 0.594 |
| New nodes test accuracy | 0.547 |
| **Node Regression Task (MSE)** | |
| Test MSE | 1.1343 |

In order to improve these results we tried to mask the number of events in the particular timestamp considering there were a lot of events occurring at a particular time stamp but we didn't observe any improvement in the results.

## 5 ABLATION STUDY

**TGN-** In this section, a detailed ablation is shown on TGN which is performing the best among other CTDG models. TGN as shown previously in Section 2 has three main modules- message, memory, and embedding. The embeddings are used in the downstream prediction task. Thus, TGN has four components - message function computation, message aggregation, embedding computation, and memory updater. Among these, the only learnable components are the embedding function and memory updater. The memory is updated with a recurrent model- GRU is used here. The embedding can be computed in several ways. Four different ways have been tested here- identity (where the feature of interaction is itself used as message), time projection (where the linear time-dependent projection of the feature is learned as

**IST-597 Project Report - Team 14**

message), sum (where message at each node is computed using a sum of the interactions between its temporal neighborhood), and attention where the sum is performed after attending over the temporal neighborhood with non-uniform weights. For the message aggregation function, two ways have been tested- mean which takes the mean of the messages at a node over its past, and last which just aggregates the current message with the last message received at the node. The number of TGN layers have also been increased to two to check if performance is enhanced. The importance of the memory module has also been tested by completely removing the modules associated with memory. Memory helps in retaining past information at a node based on its history, and the memory of each node is also referred to as the state of the node.

| Name | Embedding | Message | Message Aggr | Memory | Memory Upd | Num Layers |
|---|---|---|---|---|---|---|
| **TGN-attn** | **attention** | id | last | node | GRU | 1 |
| TGN-id | **Identity** | id | last | node | GRU | 1 |
| TGN-time | **time projection** | id | last | node | GRU | 1 |
| TGN-sum | **sum** | id | last | node | GRU | 1 |
| **TGN- mean** | attention | id | **mean** | node | GRU | 1 |
| TGN- l2 | attention | id | last | node | GRU | **2** |
| TGN- no mem | attention | - | - | - | - | 1 |

Table 5.1: Different configurations of TGN model

| | Test ap | | Test auc | | |
|---|---|---|---|---|---|
| Name | Seen | Unseen | Seen | Unseen | Time per epoch (s) |
| **TGN-attn** | **0.9858** | **0.9801** | **0.9852** | **0.9753** | ~250 |
| TGN-id | 0.9539 | 0.9297 | 0.9513 | 0.9227 | ~200 |
| TGN-time | 0.9456 | 0.9296 | 0.956 | 0.9332 | ~150 |
| TGN-sum | 0.975 | 0.9651 | 0.9731 | 0.9618 | ~200 |
| **TGN- mean** | 0.9863 | 0.9793 | 0.9854 | 0.9786 | ~500 |
| **TGN- l2** | 0.9888 | 0.9876 | 0.9883 | 0.9865 | ~1500 |
| TGN- no mem | 0.9462 | 0.9456 | 0.9359 | 0.9365 | ~150 |

Table 5.2: Performance of the different configurations of the base TGN model

The observations made based on the ablation study done on the different modules of the TGN model are listed below.

The memory component plays an important role in retaining past long-term information about the nodes, thus preventing the loss of information. It can be seen that TGN-no mem which does not have the memory module has the lowest test-AUC in comparison to the models having the memory component.

Keeping a separate embedding module takes care of the common memory staleness problem which refers to the scenario where a node has remained inactive over a long period of time, and thus its embedding used for prediction is not updated. Keeping an embedding module updates a node embedding based on its neighbors, and thus keeps getting updated even if the node itself remains inactive. Out of the different embedding modules tried, attention-based embedding is seen to have the best performance. Thus, giving more attention or importance to the more relevant temporal neighbors is crucial.

Increasing the number of TGN layers seems to slightly improve the performance of the model. But it makes the computation much more expensive than that with one layer. Thus, the TGN model needs only 1 layer for performing sufficiently well with reasonable computations. Having a memory model which keeps track of the past does not necessitate having more than one layer to capture the past information. On the other hand, models without memory like TGAT require two layers for a comparable performance with TGN with one layer. This makes the TGN model run much faster than an equally performing TGAT model. Additionally, efficient parallel processing within a batch makes TGN much more efficient.

**TGAT-** TGAT has two components- graph aggregation layer which applies attention on the spatial neighbors to compute spatial node representations, time encoder which learns temporal embeddings. The main novelty of TGAT is the temporal encoding layer. Based on the ablation study done on TGAT, the following observations are made.

| Configuration | agg_method | attn mode | time | lr | acc | auc | ap |
|---|---|---|---|---|---|---|---|
| TGAT - attn | **attn** | prod | time | 0.0001 | 0.852758495 | 0.93266872 | 0.937686821 |
| TGAT- lstm | **lstm** | prod | time | 0.0001 | 0.836378029 | 0.919634694 | 0.920088059 |
| TGAT- mean | **mean** | prod | time | 0.0001 | 0.835336743 | 0.917628124 | 0.919522877 |
| TGAT- map | attn | **map** | time | 0.0001 | 0.823436456 | 0.909887237 | 0.911662168 |
| TGAT- pos | attn | prod | **pos** | 0.0001 | 0.820112293 | 0.897993544 | 0.911662168 |
| TGAT- empty | attn | prod | **empty** | 0.0001 | 0.822122393 | 0.889844574 | 0.899055624 |
| TGAT- 0.00001 | attn | prod | time | **0.00001** | 0.818799434 | 0.878896674 | 0.887786659 |
| TGAT- 0.001 | attn | prod | time | **0.001** | 0.817899867 | 0.879667382 | 0.895776355 |
| TGAT- 0.01 | attn | prod | time | **0.01** | 0.819722909 | 0.881200954 | 0.897735283 |

Table 5.3: Different configurations of TGAT model

Unlike TGN, TGAT has no memory module to retain past information on a node. Thus, it needs at least two layers to append over the past information. This makes TGAT much more computationally expensive than TGN with 1 layer with similar performance.

Temporal encoding plays a significant role in increasing the performance of the TGAT configurations with time information incorporated. Attention mechanism used for aggregating neighbors performs significantly better than other aggregation methods like LSTM and mean. Thus, like TGN, attending over more relevant neighbors drives the performance of TGAT. The list of hyperparameters used for all the configurations of TGN and TGAT mentioned above for the accuracy reported is shown table 5.4.

| Hyperparameter | Value |
|---|---|
| Memory dimension | 172 |
| Embedding dimension | 100 |
| Time embedding dimension | 100 |
| Number of neighbours | 10 |
| Dropout | 0.1 |
| Attention heads | 2 |
| Batch size | 200 |
| Learning rate | 0.0001 |

| Hyperparameter | Value |
|---|---|
| Number of Layers | 2 |
| hidden_size | 64 |
| Dropout | 0.5 |
| Learning Rate | 0.001 |
| Weight Decay | 0.0 |
| Batch Size | 256 |
| Number of Epochs | 100 |
| patience | 10 |

Table 5.4: List of hyperparameters used in TGN (left) and TGAT (right)

IST-597 Project Report - Team 14

## 6 CHALLENGES

The major challenge was the conversion of the discrete-time dynamic graph (DTDG) to a continuous-time dynamic graph (CTDG) and making it compatible with the Event-based models. Moreover, as the area of temporal graphs is relatively new, therefore, we could get references for plenty of Node classification and link prediction tasks but a lack of resources for Node regression tasks.

## 7 FUTURE WORK

Observing these results in section 4.3 we are working on further improving the converted dataset by aggregating all the events in a particular time stamp such that we can get rid of multiple events occurring on the same node. Also, since this is a node regression task with multiple events occurring on multiple nodes (129) in a single timestamp, selecting features as the series of events for certain lags (or time steps) sounds like an interesting direction to this project and might show promising results. Agreeing with the fact that this would require computation of a larger scale.

## 8 SUMMARY

In this project, we could successfully perform a comparative analysis between the snapshot-based models namely gconv_LSTM, gconv_gru, EvolveGCNO, and EvolveGCNH, for node regression problem of disease spread prediction. These studies were carried out using the EnglandCovidDataset, a dataset for TGNs that PyTorch Geometric Temporal recently released. Using the mean squared error (MSE) measure, we assessed the performance of the models and discovered that all four models performed satisfactorily with EvolveGCN-O showing better results over others. Further, we were able to perform conversion of a DTDG dataset to the CTDG dataset and perform the analysis on the event-based model (TGN) using this converted dataset. Observations indicate that snapshot-based models tend to perform better in applications where a large number of multiple events occur at each timestamp and data is collected at regular intervals of time. CTDGs- event-based models tend to perform better in applications where very few events occur at each timestamp, and data can be collected over continuous time. Whether the EnglandCovid dataset is a suitable choice to test the CTDG models is still a question.